\documentclass[runningheads]{llncs}

\usepackage{eccv}

\usepackage{eccvabbrv}

\usepackage{graphicx}
\usepackage{booktabs}

\usepackage[accsupp]{axessibility}  

\usepackage{hyperref}

\usepackage{orcidlink}
\usepackage{bm}
\usepackage{multirow}
\usepackage{svg}

\begin{document}

\title{UniDynamics: Event-RGB Fusion for Unified Future 4D Dynamic Scene Generation} 

\titlerunning{UniDynamics}

\author{Daikun Liu \inst{1}$^\star$\orcidlink{0000-0001-6344-2517} \and
Xin Zhan \inst{2}$^\star$\orcidlink{0009-0002-2793-3524} \and
Teng Wang\inst{1}$^\dagger$\orcidlink{0000-0002-1802-0435} \and
Xiaoping Wang \inst{2}$^\dagger$\orcidlink{0000-0002-4909-8286} \and
Changyin Sun\inst{1}\orcidlink{0000-0001-9269-334X}}

\authorrunning{D.~Liu et al.}

\institute{Southeast University, China \\
\email{\{dkliu, wangteng, cysun\}@seu.edu.cn} \and
Huazhong University of Science and Technology, China \\
\email{\{xinzhan, wangxiaoping\}@hust.edu.cn}}

\maketitle
\begingroup
\renewcommand\thefootnote{}
\footnotetext{$^\star$ Equal contribution. $^\dagger$ Corresponding authors.}
\endgroup

\begin{abstract}
We propose \textbf{UniDynamics}, a diffusion-based framework for future 4D dynamic scenes (RGB, depth, and optical flow) generation from a single event-RGB pair, without requiring long histories or control priors as in existing methods, while explicitly modeling future motion fields.
The core idea is to leverage event streams to offer an alternative motion prior for single-RGB extrapolation, and to enforce geometric and motion constraints throughout generation via multimodal modeling. Specifically, we design an Event Latent Enhancement (ELE) module to align and enhance event latents into diffusion-injectable conditioning features, providing robust initial motion priors and reliable texture/structure cues. We further introduce a Perceptual Dynamics Space (PDS) embedded in the multi-scale U-Net, which decouples and adaptively interacts depth and flow while continuously feeding back constraints to appearance features, improving geometric-motion consistency for physically plausible and spatiotemporally coherent prediction. Experiments on VKitti2 and DSEC demonstrate state-of-the-art performance, producing high-quality, temporally coherent, and 4D-consistent future predictions, especially under challenging high-speed motion blur.
  
\end{abstract}

\section{Introduction}
\label{sec:intro}
In autonomous driving scenarios, the ability of infering the evolution of the environment over time and understand future scenes based on current visual observations is critical for autonomous systems. It serves as an essential foundation for motion planning~\cite{planning}, obstacle avoidance~\cite{avoidance}, and risk-aware decision~\cite{risk}.

\begin{figure}[t]
    \centering
    \includegraphics[width=1.0\textwidth,height=0.22\textwidth]{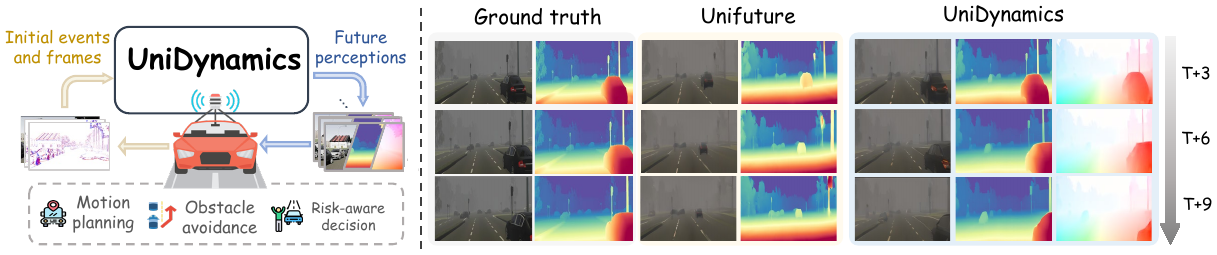}
    \captionof{figure}{Our \textbf{UniDynamics} infers future 4D scenes using only the current event-RGB pair. Leveraging motion cues and texture priors of events, it jointly models appearance, geometry, and motion, producing more faithful and spatiotemporally coherent predicts, especially under motion blur, where UniFuture~\cite{unifuture} with only a blurred RGB fails.}
    \label{fig_1}
\end{figure}

Some existing works focus on generative driving world models~\cite{DWM_vista, DWM_dreamer, DWM_GEM, unifuture, DWM_GAIA1AG, drivedreamer2}, which leverage the current frame or historical observations and introduce conditions such as trajectories, velocities, or text to improve controllability of future generations. However, the reliance on additional conditioning signals substantially increases system complexity and adaptation cost. In contrast, vision-only video prediction methods~\cite{karypidis2025dinoforesight} model the temporal evolution of semantic features from pretrained video foundation models, better matching the need for a ``plug-and-play'' perception front-end, but their fine-grained fidelity is often inferior to generative approaches. More importantly, in real-world open-road deployment, long video histories and consistent structured priors are not always reliably available, making a more practical and transferable setting that uses only a single current frame as the minimal input. Recent Unifuture~\cite{unifuture} extrapolates future scenes under single-frame conditioning by leveraging motion-prior conditions (e.g., velocity, acceleration, and trajectory). However, without such priors, scene dynamics are unobservable, and motion blur further degrades appearance cues, leading to erroneous extrapolation (see UniFuture in~\cref{fig_1}).

Notably, the inherent temporal sensitivity of event cameras~\cite{DVS1, DVS2} provides directional and boundary motion cues, offering an alternative motion prior for future extrapolation under single-frame conditioning. In particular, event streams offer continuous motion observations without requiring long histories, and remain informative even under fast-motion blur, providing stable initial momentum and structural cues. \textit{We therefore argue that event streams provide a more stable and blur-robust motion prior, enabling reliable future extrapolation even under single-frame conditioning.} In addition, the above methods often restrict prediction to static RGB appearance~\cite{DWM_vista, DWM_dreamer, DWM_GAIA1AG} or depth geometry~\cite{unifuture, DWM_GEM, karypidis2025dinoforesight} and lack explicit motion-field modeling, which is crucial for assessing scene dynamics, interaction risks, and potential collisions~\cite{flowforavoidance}, and serves as a key constraint for physically consistent and temporally coherent future-scene prediction and percption.

To this end, we propose \textbf{UniDynamics}, a diffusion-based RGB, depth, and optical flow predicting framework with an event-RGB pair observation, as illustrated in~\cref{fig_1}. The core of our approach lies in event-guided condition enhancement and multimodal jointly interacting and predicting. Specifically, we treat the event stream as complementary to the RGB observation and introduce an Event Latent Enhancement (ELE) module to encode it into the event latent. To obtain stable and generalizable event cues, we use the RGB latent as an appearance reference and align/enhance the event latent via cross-attention layers, yielding a condition that provides initial momentum and texture/boundary cues while remaining compatible with RGB-pretrained SVD~\cite{svd}. Since appearance, geometry, and motion are heterogeneous views of the same scene, we adopt the latent space of a pretrained image autoencoder as a shared alignment domain, diffuse on the image latent, and derive depth and flow from it, avoiding modality-specific encoders and improving cross-modal alignment. We further embed a Perceptual Dynamics Space (PDS) into the multi-scale U-Net to enable continuous interaction between geometry/motion priors and appearance features. Task-specific adapters and gated interactions in PDS decouple depth and flow to reduce multi-task interference and suppress flow-induced structural perturbations, improving temporal coherence while preserving stable structure.

With event-based motion encoding and the proposed PDS-driven multimodal interaction and refinement, UniDynamics achieves high-fidelity and spatiotemporally coherent future 4D scene predicting, particularly in challenging scenarios with high-speed motion blur, as illustrated in~\cref{fig_1}. Experiments on the synthesized event-based VKitti2~\cite{VKitti2} and the zero-shot generalization studies on DSEC~\cite{dsec} demonstrate that our method achieves higher-quality and more spatiotemporally coherent future scene generation and perception. Our main contributions are summarized as follows:
\begin{itemize}
 \item We propose \textbf{UniDynamics}, a diffusion-based future 4D scene generation framework that fuses an event stream with an RGB observation, enabling physically plausible and spatiotemporally consistent modeling of future scene appearance (RGB), depth, and optical flow.
 \item We introduce an event latent enhancement module that yields generalizable and diffusion-injectable event conditions, providing stable initial motion cues and rich texture/boundary information under an RGB input.
 \item We design a Perceptual Dynamics Space integrated into the U-Net, where depth and optical flow are decoupled and modeled in a latent representation space aligned with RGB. Through feedback injection, PDS enables bidirectional interaction with appearance latent, improving structural stability and spatiotemporal coherence.
 \item Our method delivers higher-quality and more spatiotemporally consistent future 4D scene prediction and perception on VKitti2 and DSEC, particularly in challenging scenarios such as high-speed motion blur.
\end{itemize}

\section{Related Work}
\noindent\textbf{RGB-based Future Scene Prediction Methods.} 
Future scene prediction from visual observations is critical for autonomous driving. Generic video prediction methods~\cite{predrnn, SimVP, TAU} are mostly developed on general datasets and often transfer poorly to driving scenarios. Driving-focused generative world models~\cite{DWM_GAIA1AG, DWM_dreamer, DWM_vista, DWM_think, drivedreamer2, ADriver-I} improve controllability by introducing additional conditions such as actions, text, trajectories, and other priors, and some further model RGB with depth~\cite{unifuture, DWM_GEM}. However, the reliance on extra structured conditions increases system complexity and adaptation cost. Vision-only prediction methods based on CNN/RNN, masked modeling, or foundation models~\cite{PFA, FLODCAST, FUTURIST, karypidis2025dinoforesight} avoid extra priors, but typically have lower fine-grained fidelity and often require long histories, incurring memory and latency. However, in deployment-relevant single-frame settings, scene dynamics are difficult to capture, and appearance cues are often degraded under fast motion. Meanwhile, most methods lack explicit motion-field modeling, limiting effective spatiotemporal consistency constraints. Motivated by these gaps, we introduce event streams as dynamic observations for the deployment-relevant settings, and adopt a diffusion framework to jointly predict RGB, depth, and optical flow for physically plausible and spatiotemporally coherent 4D scene generation and perception.

\noindent\textbf{Event-based Future Scene Prediction Methods.}
Compared to conventional cameras, event cameras provide reliable motion observations for predicting in dynamic scenes. Recent works~\cite{emotion, DESSERT, das2025fastfeaturefield} validate the effectiveness of events for temporal extrapolation. E-Motion~\cite{emotion} combines the high temporal resolution of events with video diffusion to predict future event frames, while $F^3$~\cite{das2025fastfeaturefield} learns predictive representations by predicting events and transferring them to tasks such as optical flow and semantic segmentation. However, event-only methods often break down in static or low-texture regions due to the dynamic characteristics and sparsity of events. DESSERT~\cite{DESSERT} fuses events with RGB for diffusion-based single-frame prediction, but long-horizon prediction suffers from error accumulation. Moreover, these methods also lack explicit motion-field modeling. Therefore, we exploit the complementarity of events and RGB and jointly model 4D scenes to achieve temporally coherent future evolution.

\noindent \textbf{Event-RGB-fusion-based Vsion Tasks}
Monocular depth and optical flow estimation are core vision tasks for autonomous driving. Since events and frames offer complementary temporal and appearance cues, many works fuse them to improve robustness and accuracy.
Early works on event–RGB depth estimation make efforts on using recurrent fusion~\cite{RAMnet}, flow-based compensation~\cite{EVEN}, and cross-modal consistency with pseudo-supervision~\cite{EMoDepth}. Later methods improve fusion robustness via structural priors~\cite{SRFNet}, high-level pattern/token interactions~\cite{PCDepth}, cross-modal attention with multi-scale interaction~\cite{ER-F2D, UniCT}, or frequency-domain fusion~\cite{FUSE}. For optical flow, several approaches extend RAFT~\cite{raft} to fuse events and RGB, while DCEIFlow~\cite{DCEIFlow} and BFlow~\cite{BFlow} estimate dense continuous-time motion, and STFlow~\cite{STFlow} narrows the modality gap with spatially guided temporal aggregation. Diffusion-based fusion~\cite{E-FlowDiffuser} is also explored by incorporating events into FlowDiffuser~\cite{flowdiffuser} for challenging conditions. Although these methods target current/past perception rather than prediction, they consistently demonstrate the complementarity of RGB–event fusion, motivating our event–RGB scheme for future scene prediction.

\begin{figure*}[t]
\centering
\includegraphics[width=1.0\textwidth,height=0.37\textwidth]{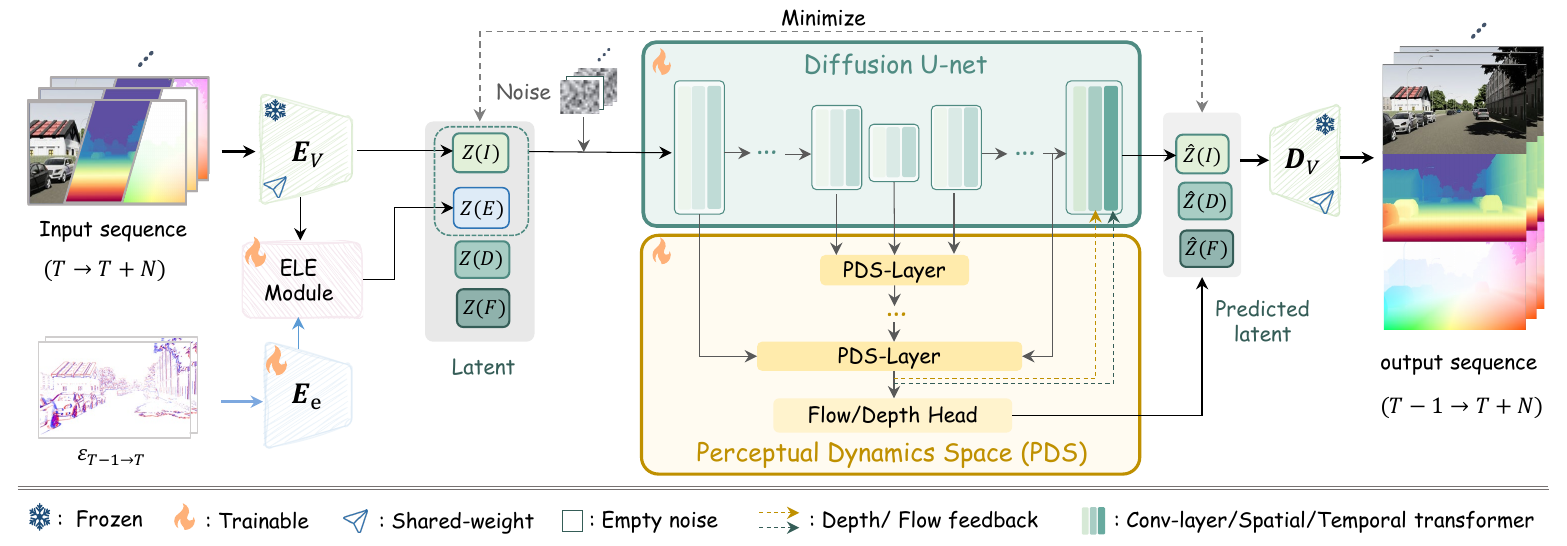}
\caption{The training pipeline of UniDynamics. The event stream is encoded by the Event Latent Enhancement (ELE) module into diffusion-injectable features that provide initial motion and texture cues. RGB, depth, and optical flow are derived from the same RGB latent: RGB is generated via diffusion denoising, while depth and flow are jointly learned through an additional Perceptual Dynamics Space (PDS). These modalities interact and are refined continuously across multiple scales, enabling physically plausible and spatiotemporally coherent future scene prediction.
}
\label{fig_2}
\end{figure*}

\section{UniDynamics}
\subsection{Preliminaries}
Stable Video Diffusion (SVD)~\cite{svd} generates high-fidelity videos via latent diffusion. It encodes frames into a VAE latent space, progressively adds Gaussian noise to the target latent sequence during training, and optimizes a spatiotemporal U-Net to learn the reverse denoising process conditioned on the input-frame latent. At inference, it starts from Gaussian noise, iteratively denoises under the same conditioning, and decodes the resulting latents into video frames. Leveraging high-fidelity generation of SVD, recent works~\cite{DWM_dreamer, DWM_vista, unifuture, DWM_GEM} synthesize realistic future driving scenes and even extend to geometry-aware prediction~\cite{unifuture, DWM_GEM}. However, under the deployment-relevant single-frame setting, where initial motion cues cannot be provided, future extrapolation becomes highly ill-posed, and limited dynamics modeling further restricts temporal coherence.

\subsection{Overview}
We propose \textbf{UniDynamics}, a diffusion-based framework for future 4D scene generation, as illustrated in~\cref{fig_2}. Given the current frame at time $T$ and corresponding event stream $\mathcal{E}_{T-1 \to T}$, we build on SVD as the generative backbone and aim to predict the future $N$-frame RGB sequence along with the depth maps and optical flows. Since RGB, depth, and optical flow provide complementary views of the same scene, we adopt the latent space of a pretrained image VAE encoder $\bm{E}_V$ as a unified representation. The diffusion backbone generates future appearance only in the image latent $Z(I)$, while depth and flow latents are initialized from $Z(I)$ and progressively refined through multi-scale U-Net interactions, enabling joint modeling of appearance, geometry, and motion in a shared latent space. This avoids training extra encoders, promotes cross-modal alignment, and allows lightweight networks to translate between modalities. To provide initial momentum under single-frame conditioning and support joint multimodal generation, our architecture comprises two key components. \textbf{(1) Event latent guided motion conditioning.} We extract motion cues from the event stream and employ an Event Latent Enhancement module to produce a diffusion-injectable conditioning feature $Z(E)$, providing stable motion signals and rich texture/boundary priors that complement RGB. \textbf{(2) Perceptual Dynamics Space for depth and flow modeling.} To capture the coupled evolution of appearance, geometry, and motion in future scenes, we introduce the Perceptual Dynamics Space (PDS), a multi-scale interaction module embedded within the U-Net. PDS jointly models depth and optical-flow latents and performs bidirectional feature exchange with the diffusion backbone to impose multimodal consistency and refine the generated predictions. Finally, the predicted latent codes $\hat{Z}(I)$, $\hat{Z}(D)$, and $\hat{Z}(F)$ are decoded by the shared VAE decoder $\bm{D}_V$ into RGB frames, depth maps, and optical flow, enabling joint generation of future 4D scenes.

\subsection{Event Latent Guided Motion Injection}
\noindent \textbf{Event Representation.}
Event cameras generate event stream $\mathcal{E}_{T-1 \to T}$ with $N_e$ events $\{x_i, y_i, t_i, p_i\}_{i=0}^{N_e}$ based on brightness changes, where $(x,y)$, $t$, and $p$ are coordinates, timestamp, and polarity, respectively. To extract motion cues, we convert events into voxel grids with $B$ temporal bins~\cite{zhu_unflow}:
\begin{align}
\mathcal{V}\left(b, x, y\right)&=\sum_{i=0}^{N_e-1} p_{i}\,\delta(x-x_i)\,\delta(y-y_i)\,\max \left(0,1-\left|b-t_{i}^{*}\right|\right),\label{eq1}\\
t_{i}^{*} &= \frac {(t_{i} - t_{min})}{t_{max} - t_{min}} \times (B - 1), \label{eq2}
\end{align}
where $\delta(\cdot)$ denotes the Kronecker delta function, $b \in [0, B-1]$ is the index of the temporal bin, and $\max(0, 1-|b-t_i^*|)$ represents the linear interpolation kernel along the temporal dimension.

\begin{figure*}[t]
\centering
\includegraphics[width=1.0\textwidth,height=0.36\textwidth]{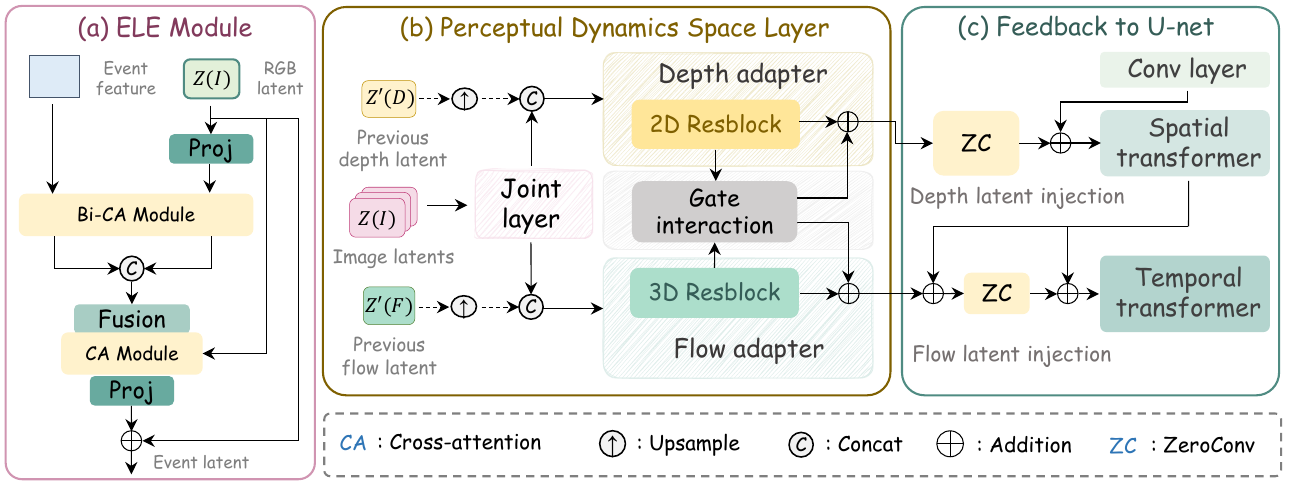}
\caption{UniDynamics submodules: (a) ELE module, which exploits the complementarity of events and RGB to produce an event latent with effective initial momentum and rich texture cues; (b) PDS layer, which decouples the depth and flow spaces to enable independent learning with adaptive interaction; and (c) depth geometry and flow motion cues are injected back into the U-Net’s spatial and temporal layers, respectively, providing effective constraints on RGB evolution.}
\label{fig_4}
\end{figure*}

\noindent \textbf{Event Latent Enhancement.}
We encode event frames into event features $F_e$ using a learnable event encoder $\bm{E}_e$ comprised of 2D resblocks. To obtain a stable event latent that provides complementary cues for single RGB, we introduce an Event Latent Enhancement (ELE) module, as shown in~\cref{fig_4} (a). 
Specifically, we first align event features with the RGB latent using bidirectional cross-attention: 
\begin{align}
Z'(E) &= {\rm BiCA}((F_e, {\rm Proj}(Z(I))))+ F_e,\label{eq3}\\
Z'(I) &= {\rm BiCA}(({\rm Proj}(Z(I)), F_e))+ {\rm Proj}(Z(I)), \label{eq4}
\end{align}
where $\mathrm{Proj}(\cdot)$ denotes a linear projection, and the event and RGB features are alternately used as the queries, keys, and values for each other. We then perform modality fusion to $Z'(E)$ and $Z'(I)$ with convolution and apply cross-modal attention with the RGB latent as key and value to leverage RGB appearance information for enhancing the final event latent $Z(E)$:
\begin{equation}\label{eq5}
Z(E) = {\rm Proj}({\rm CA}({\rm Fusion}(Z'(E), Z'(I)), Z(I)))) + Z(I).
\end{equation}
This design enriches the event latent with rich appearance of RGB, especially when events are sparse, while ensuring latent-space compatibility, leading to more effective and robust conditioning for the diffusion process.

Since the event latent primarily encodes short-term motion, directly concatenating $Z(E)$ with future noise latents as a strong condition for all future steps (as done for RGB latents) may overly entangle transient cues with global scene evolution, introducing bias and causing structural artifacts, temporal drift, or long-horizon overfitting. Instead, we adopt an SVD-compatible, softer conditioning scheme: We treat $Z(E)$ as the latent of a ``historical-state RGB'' at $T-1$, add noise to it jointly with the future latent sequence during training, and concatenate the current RGB latent $Z(I)$ to each noisy latent along the channel dimension as the reference. The model then generates frames at $\{T-1,\ldots, T+N\}$. This formulation embeds events as an implicit initial-momentum prior rather than a hard constraint, allowing the spatiotemporal U-Net to adaptively extract and propagate motion cues through multi-scale aggregation, while also providing useful texture/structure guidance when RGB is blurred.

\subsection{Perceptual Dynamics Space for Depth-Flow Modeling}
We embed the multi-scale interaction space into four stages $\{1, \frac{1}{2}, \frac{1}{4}\}$ of the U-Net, where two task-specific adapters decouple the unified latent representation into geometry and motion subspaces to model the evolution of depth and flow. A gated bidirectional interaction further regulates information exchange between them, and the resulting geometric and motion cues are injected into the spatial and temporal layers of the U-Net decoder, respectively, to regularize RGB generation and improve structural stability and spatiotemporal coherence.

\noindent \textbf{Depth-Flow Adapter and Gated Bi-interaction.}
As shown in~\cref{fig_4} (b), each PDS layer first fuses the features from the corresponding U-Net encoder and decoder blocks, together with the output of the previous PDS layer, using a joint layer with linear projection. The fused features are then decoupled, via feature reuse, into depth and flow adapters to alleviate gradient ``tug-of-war'' in multi-task learning. In addition, each adapter also takes the outputs $Z'(D)$ and $Z'(F)$ from the previous PDS layer as input, enabling continuous cross-layer modeling. The depth adapter uses 2D convolutions to capture spatial structure and outputs $F_d$, while the flow adapter uses 3D convolutions to model spatiotemporal dependencies and outputs $F_f$.

Depth and flow are tightly coupled: motion boundaries often align with structural edges, and motion patterns are strongly constrained by scene geometry. To capture this complementarity, we introduce a gated bidirectional interaction module in each PDS layer to enable adaptive information exchange between the depth and flow subspaces. Concretely, we first concatenate them to form a fused representation $F_{fu}$. Then we use two convolutional branches to generate cross-modal updates, whose injection strengths are adaptively modulated by sigmoid gates and added to the residuals:
\begin{align}
\bar{Z}(D) &= {\rm Sigmoid}({\rm Conv_f}(F_{fu}))\ast F_f + F_d,\label{eq7}\\
\bar{Z}(F) &= {\rm Sigmoid}({\rm Conv_d}(F_{fu}))\ast F_d + F_f, \label{eq8}
\end{align}
The resulting features are then injected back into the deocoder of U-Net, and also upsampled and passed to the next PDS layer. After multiple layers, we obtain the final depth and flow latent representations $\hat{Z}(D)$ and $\hat{Z}(F)$.

\noindent \textbf{Depth and Flow Latent Injection.}
To impose geometric and motion constraints throughout diffusion generation, we feed back the depth/flow latents modeled in PDS into the decoder of the U-Net backbone, as shown in~\cref{fig_4} (c). Since depth captures static geometry and boundaries that directly regularize spatial details, while optical flow describes inter-frame motion and occlusion evolution that better guides temporal aggregation and dynamics propagation, we inject depth and flow latents $X_d$ and $X_f$ into the U-Net’s spatial and temporal layers, respectively. To avoid overly strong constraints early in the training stage that could disrupt pretrained priors, we adopt a zero-initialized convolutional injection strategy for both depth and flow:
\begin{equation}\label{eq9}
Y_d = {\rm ZeroConv}(X_d), Y_f = {\rm ZeroConv}(X_f + X_t),
\end{equation}
where $X_t$ is the temporal feature in the current U-Net block. $Y_d$ and $Y_f$ are then added to the spatial and temporal features, respectively:
\begin{equation}\label{eq10}
X'_s = Y_d + X_s, X'_t = Y_f + X_t,
\end{equation}
where $X_s$ is the spatial feature of the U-Net block. By initializing the weights of ZeroConv to zero, the network starts with an equivalent of ``no injection” and can stably learn to adaptively increase the injection strength during training, achieving a progressive fusion of constraints from weak to strong. 

\begin{figure}[t]
\centering
\includegraphics[width=0.9\textwidth,height=0.25\textwidth]{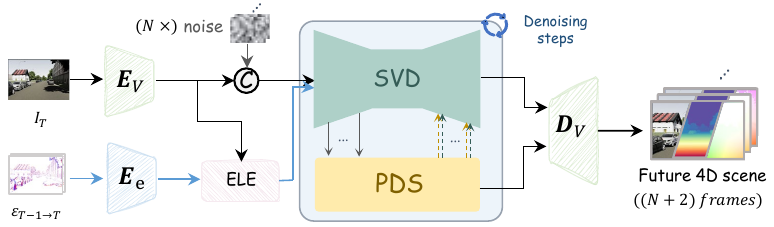}
\caption{UniDynamics inference pipeline: given a single event–RGB pair, it simultaneously predicts future RGB, depth, and optical flow. The RGB latent is concatenated with noise and, together with the event latent, fed into the U-Net for denoising to obtain multimodal latents, which are finally decoded by a shared decoder into future RGB frames, depth maps, and optical flow.}
\label{fig_3}
\end{figure}

\subsection{Training and Inference}
Our framework forecasts the future $N$ frames together with frames at $T-1$ and $T$ of RGB, depth, and optical flow conditioned on a paired of event-RGB. We use a weight-shared VAE encoder to map multiple modalities into latent representations, and perform unified modeling on the shared RGB latent space to obtain $\hat{Z}(I)$, $\hat{Z}(D)$, and $\hat{Z}(F)$, which are decoded by the same VAE decoder to produce 4D scene predictions. 

As shown in~\cref{fig_2}, during training, we add noise to the non-conditioning frames and the event-enhanced latent frame, concatenate the conditioning-frame latent, and jointly generate the pseudo-history, current, and future RGB-depth-flow outputs. We optimize generation and perception tasks with a weighted combination of losses. Specifically, for RGB, we compute a latent-space loss $L(I)$ between $Z(I)$ and $\hat{Z}(I)$, including an MSE term, a dynamics-enhanced term, and a structure-preserving term. For depth, we use an analogous loss $L(D)$, and additionally apply a scale- and shift-invariant loss $L_{\text{SSI}}$ in pixel space between the predicted and target depth~\cite{Ranftl2022}. For optical flow, the loss consists of a latent-space MSE term $L(F)$ and a pixel-space $L_1$ term. We additionally apply an MSE loss $L(E)$ between $Z(E)$ and the RGB-frame latent at $T-1$ to stabilize training of the ELE module. The overall objective is:
\begin{equation}\label{eq11}
L = L(I) + L(D) + L(F) + L(E) + \lambda_1 L_{\text{SSI}} + \lambda_2 L_1,
\end{equation}
where $\lambda_1$, $\lambda_2$ are the balance coefficient.

As the inference pipeline in~\cref{fig_3}, we concatenate the current RGB latent with the noise-free event-enhanced latent and $N$ Gaussian noise maps, and feed them into the U-Net backbone for iterative denoising. Together with PDS-based joint modeling and multimodal interaction, this enables physically plausible and spatiotemporally coherent future scene prediction and perception.

\section{Experiments}
\subsection{Datasets and Metrics}
\noindent \textbf{Datasets.} We train and evaluate on the driving dataset E-VKitti2~\cite{VKitti2}, and conduct zero-shot testing on real-world DSEC~\cite{dsec}, MVSEC~\cite{mvsec}, and M3ED~\cite{M3ED} datasets to assess generalization. VKitti2 provides high-quality synthetic RGB images, depth maps, and optical flow across diverse driving scenarios and appearance conditions (e.g., varying illumination). We use v2e~\cite{v2e} to synthesize E-VKitti2 dataset. DSEC is a real-world event-camera dataset containing RGB frames, event streams, disparity, and optical flow under different lighting and dynamic scenes. We convert the official disparity maps to depth using the provided conversion code. The $outdoor\_day1$ sequence of the MVSEC provides relatively sparse events, enabling evaluation of generalization under weak event cues. Moreover, the M3ED features complex and dynamic outdoor scenes. To further evaluate performance under high-speed motion, we apply flow-based motion blur on E-VKitti2 to emulate image blur, where the prediction target is the corresponding deblured RGB.

\noindent \textbf{Metrics.}
We evaluate RGB generation using Fr\'{e}chet Inception Distance (FID) and Fr\'{e}chet Video Distance (FVD) to assess visual quality and spatiotemporal coherence. For depth, we report absolute relative error (AbsRel), accuracy $\delta_n$ ($\delta < 1.25^n$, $n=1,2,3$), and root mean squared error  (RMSE) to measure accuracy and structural consistency. For optical flow, we use endpoint error (EPE), angular error (AE), and the percentage of pixels with EPE larger than $n$ pixels ($n$PE, $n=1,2,3$) to quantify accuracy and robustness.

\subsection{Implementation details}
For event representations, we use a 100 ms event window and split into $B$=15 temporal bins. We crop and resize inputs to $192\times640$ for E-VKitti2 and $320\times384$ for DSEC, MVSEC, and M3ED. We train the model on E-VKitti2 with the AdamW optimizer using an initial learning rate of $5\times10^{-5}$, batch size 1, on 2$\times$A100 GPUs for 16k steps. In the multi-task loss, we set $\lambda_1=2.0$ and $\lambda_2=1.0$. To stabilize training, we apply an exponential moving average (EMA) to smooth model weights and adopt classifier-free guidance by randomly dropping the conditioning branch with probability 15\% during training. For memory efficiency, we use DeepSpeed ZeRO-2~\cite{deepspeed} to reduce optimizer-state and gradient memory. At the evaluation phase, conditioned on a single frame and events, we predict future $N=9$ RGB frames along with the corresponding depth and optical flow, and compare them with ground truth. More details in the supplementary material.

\subsection{Main Results}
\noindent \textbf{Results on E-VKitti2.}
\cref{tab:full_comparison} reports the results on the E-VKitti2 dataset. The compared methods include our proposed UniDynamics and its variants, non-diffusion-based methods, and diffusion-based methods. For a fair evaluation, we retrain SimVP~\cite{SimVP}, TAU~\cite{TAU}, FLODCAST~\cite{FLODCAST}, FUTURIST~\cite{FUTURIST}, SVD~\cite{svd}, Vista~\cite{DWM_vista}, and UniFuture~\cite{unifuture} under the same dataset settings and evaluation protocol. Specifically, \textit{SVD-Dpeth} and \textit{SVD-Flow} denote the SVD-based only depth
and flow generation model, respectively. \textit{UniFuture-Depth} and \textit{UniFuture-Flow} correspond to the original RGB-depth generation and RGB-flow generation settings, respectively. For ours, we report the \textbf{UniDynamics} and an ablation \textit{UniDynamics (w/o events)} to verify the contribution of event inputs.

\begin{table*}[t] 
\caption{Quantitative comparison on E-VKitti2~\cite{VKitti2}. ``$\uparrow$'' and ``$\downarrow$'' indicate that higher and lower values are better, respectively. Best results are highlighted in bold. For unsupported tasks, we mark them as ``Unsupported''.
}
\label{tab:full_comparison}
\centering
\resizebox{1.0\linewidth}{!}{
\begin{tabular}{lcccccccccccc}
\toprule
\multirow{2}{*}{Method} & 
\multicolumn{2}{c}{Generation} & 
\multicolumn{5}{c}{Depth} & 
\multicolumn{5}{c}{Optical Flow} \\
\cmidrule(lr){2-3} \cmidrule(lr){4-8} \cmidrule(lr){9-13}
 &
 FID $\downarrow$ & FVD $\downarrow$ & 
 AbsRel $\downarrow$ & $\delta_1 \uparrow$ & $\delta_2 \uparrow$ & $\delta_3 \uparrow$ & RMSE $\downarrow$ & 
 EPE $\downarrow$ & AE $\downarrow$ & 1PE $\downarrow$ & 2PE $\downarrow$ & 3PE $\downarrow$ \\
\midrule

\multicolumn{13}{c}{\textit{Only future generation}} \\
\midrule
SimVP~\cite{SimVP} & 137.8 & 1301.7 & \multicolumn{10}{c}{Unsupported} \\
TAU~\cite{TAU} & 127.8 & 1530.4 & \multicolumn{10}{c}{Unsupported} \\
SVD~\cite{svd}  & 114.5 & 1244.4 & \multicolumn{10}{c}{Unsupported} \\
Vista~\cite{DWM_vista} & 101.8 & 1364.6 & \multicolumn{10}{c}{Unsupported} \\
\midrule
\multicolumn{13}{c}{\textit{Only future perception}} \\
\midrule
FLODCAST~\cite{FLODCAST} & \multicolumn{2}{c}{Unsupported} & 0.57 & 21.4 & 52.6 & 70.1 & 28.8 & 11.7 & 55.7 & 76.4 & 63.7 & 52.4 \\
FUTURIST~\cite{FUTURIST} & \multicolumn{2}{c}{Unsupported} & 0.47 & 29.7 & 67.7 & 84.9 & 44.9 & \multicolumn{5}{c}{Unsupported} \\
 SVD-Depth~\cite{svd} & \multicolumn{2}{c}{Unsupported} & 0.39 & 52.8 & 84.1 & 90.8 & 61.4 & \multicolumn{5}{c}{Unsupported} \\
 SVD-Flow~\cite{svd} & \multicolumn{2}{c}{Unsupported} & \multicolumn{5}{c}{Unsupported} & 8.5 & 40.3 & 88.8 & 74.1 & 63.3 \\
\midrule

\multicolumn{13}{c}{\textit{Joint generation and single perception}} \\
\midrule
UniFuture-Depth~\cite{unifuture} & 44.0 & 584.1 & 0.38 & 56.8 & 83.9 & 90.1 & 59.4 & \multicolumn{5}{c}{Unsupported} \\

UniFuture-Flow~\cite{unifuture} & 46.9 & 615.3 & \multicolumn{5}{c}{Unsupported} & 5.9 & 33.8 & 78.8 & 63.1 & 52.1 \\

\midrule
\multicolumn{13}{c}{\textit{Unify future perception and generation}} \\
\midrule
UniDynamics (w/o events) & 46.8 & 577.1 & 0.36 & 56.0 & 83.9 & 90.3 & 59.0 & 5.3 & 34.1 & 79.2 & 61.5 & 49.1 \\
{UniDynamics (w/o RGB)} & 79.2 & {1048.4} & {0.50} & {34.3} & {72.8} & {86.1} & {71.7} & {3.4} & {18.6} & {57.5} & {41.5} & {31.6} \\
\midrule
\textbf{UniDynamics} (ours) & \textbf{39.1} & \textbf{223.6} & 
\textbf{0.27} & \textbf{67.1} & \textbf{87.8} & \textbf{92.7} & \textbf{54.7} & 
\textbf{3.0} & \textbf{21.0} & \textbf{58.1} & \textbf{38.5} & \textbf{27.7}\\
\bottomrule
\end{tabular}
}
\end{table*}

The results show that only our method supports the simultaneous generation of future RGB, depth, and optical flow. Moreover, diffusion-based methods show overall superiority over non-diffusion-based methods in terms of both video generation and future perception. Comparing \textit{Vista} with \textit{UniDynamics (w/o events)} verifies that additionally modeling depth and flow to guide RGB synthesis significantly improves generation quality (from 101.8 to 46.8 on FID) and spatiotemporal coherence (from 1364.6 to 577.1 on FVD). Meanwhile, the results of \textit{UniDynamics (w/o RGB)} show that event-only prediction can still outperform Vista. However, due to the absence of rich appearance references, it yields lower generation quality and perception accuracy than the full UniDynamics model. Compared with \textit{UniFuture} variants and \textit{UniDynamics (w/o events)}, our method achieves slightly better FID, delivers a clear gain in RGB temporal coherence, and also yields stronger performance on the perception tasks. These gains stem from leveraging events to provide initial motion cues and rich texture/structure information, together with the PDS module that enables continuous depth–flow–RGB interaction. Overall, our method delivers higher-fidelity (the lowest FID of 39.1) and more coherent RGB generates (the lowest FVD of 223.6), more accurate and structurally consistent depth, and lower-error, more robust optical flow, demonstrating the effectiveness of event-guided multimodal modeling for future scene generation and perception.

\begin{figure*}[t]
    \centering    \includegraphics[width=1.0\textwidth,height=0.58\textwidth]{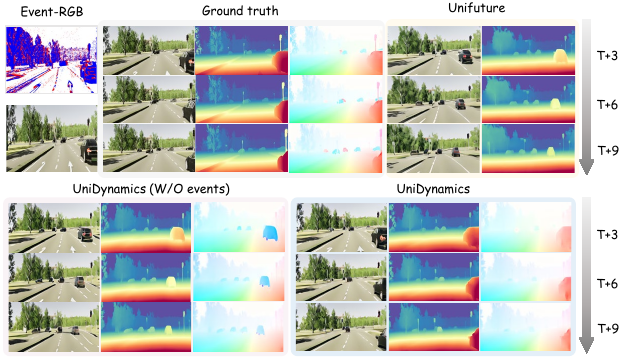}
    \captionof{figure}{Qualitative comparisons of future 4D scene prediction on E-VKitti2. UniDynamics preserves spatiotemporal coherence and yields more accurate perception results.}
    \label{main_viz}
\end{figure*}

\begin{figure*}[t]
    \centering    \includegraphics[width=1.0\textwidth,height=0.4\textwidth]{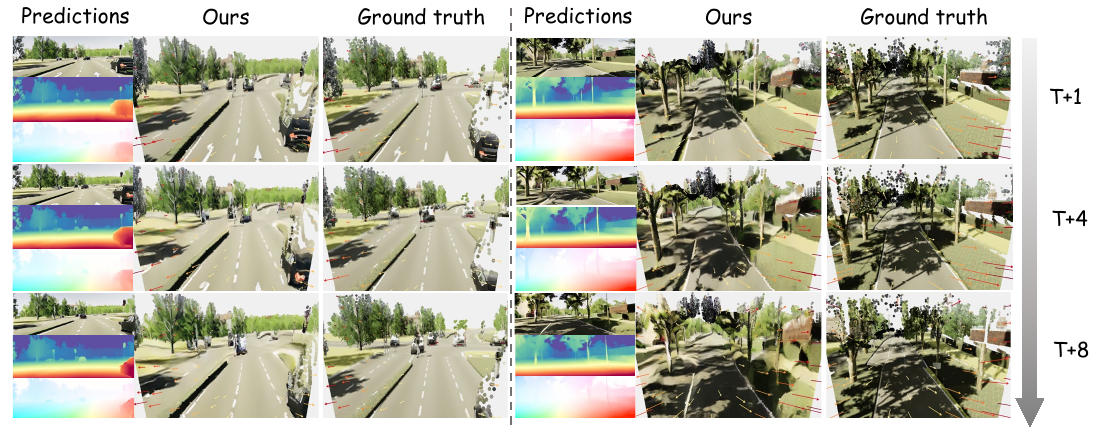}
    \captionof{figure}{The reconstructed 3D point cloud and scene flow, obtained from the future RGB, depth, and optical flow, demonstrate our ability to predict future 4D scenes.}
    \label{4Drecon}
\end{figure*}

As shown in~\cref{main_viz}, \textbf{UniDynamics} generates RGB frames that better follow realistic scene evolution, leading to more accurate depth and flow predictions that capture scene geometry and motion more faithfully. In contrast, the baselines produce RGB frames with incorrect motion, resulting in less accurate depth and optical flow. These results demonstrate the effectiveness of our event-guided joint RGB–depth–flow modeling for coherent future forecasting and perception. Moreover, \cref{4Drecon} visualizes the 3D point clouds reconstructed from the predicted RGB and depth, as well as the scene flow derived from the predicted optical flow. The reconstructions and scene flow closely match the ground truth in overall structure and motion trends, indicating that \textbf{UniDynamics} has the potential to enable 4D scene reconstruction of future worlds. More visualizations are provided in the supplementary material.

\noindent \textbf{Robustness under Motion Blur.}
We evaluate robustness to motion blur on E-VKitti2 by simulating image degradation under high-speed motion. ~\cref{tab:motion_blur_robustness} compares our method with the strongest baseline \textit{UniFuture} and our ablation \textit{UniDynamics (w/o events)}. With a single blurred RGB input, the UniFuture family and \textit{UniDynamics (w/o events)} perform poorly on all generation and perception metrics, consistent with the qualitative comparison in~\cref{fig_1}. Such challenging scenarios lack a reliable appearance reference, which degrades generation quality as well as depth/flow estimation accuracy. These results indicate that event signals can provide effective texture and structural cues to compensate for blurred RGB in high-speed scenes, and that multimodal joint modeling is crucial for robust future prediction.

\begin{table*}[t]
\caption{Performance comparison under \textbf{motion blur} conditions on the E-VKitti2~\cite{VKitti2}.}
\label{tab:motion_blur_robustness}
\centering
\resizebox{1.0\linewidth}{!}{
\begin{tabular}{lccccccccccccc}
\toprule
\multirow{2}{*}{Method} & \multicolumn{2}{c}{Generation} & \multicolumn{5}{c}{Depth} & \multicolumn{5}{c}{Optical Flow} \\
\cmidrule(lr){2-3} \cmidrule(lr){4-8} \cmidrule(lr){9-13}
 & FID $\downarrow$ & FVD $\downarrow$ & 
 AbsRel $\downarrow$ & $\delta_1 \uparrow$ & $\delta_2 \uparrow$ & $\delta_3 \uparrow$ & RMSE $\downarrow$ & 
 EPE $\downarrow$ & AE $\downarrow$ & 1PE $\downarrow$ & 2PE $\downarrow$ & 3PE $\downarrow$ \\



\midrule
{UniFuture-Flow}~\cite{unifuture} & 61.4 & {644.3} & \multicolumn{5}{c}{Unsupported} & {5.67} & {32.3} & {78.4} & {62.6} & {51.4} \\

{UniFuture-Depth}~\cite{unifuture} & 56.1 & {668.7} & {0.391} & {57.9} & {84.1} & {90.1} & {59.5} & \multicolumn{5}{c}{Unsupported} \\

{UniDynamics (w/o events)} & 59.8 & {617.7} & {0.288} & {50.2} & {79.9} & {84.2} & {58.5} & {5.40} & {33.8} & {78.6} & {60.9} & {48.7} \\

\textbf{UniDynamics} (ours) & \textbf{53.6} & \textbf{272.8} & \textbf{0.265} & \textbf{68.1} & \textbf{88.0} & \textbf{92.9} & \textbf{54.4} & \textbf{2.86} & \textbf{19.5} & \textbf{55.4} & \textbf{36.0} & \textbf{25.9}\\
\bottomrule
\end{tabular}
}
\end{table*}

\begin{table*}[t]
\caption{Zero-shot results on the real-world datasets. ``-'' indicates that the result is either not reported or the dataset does not support the computation of the metric.}
\label{tab:dsec_zeroshot}
\centering
\resizebox{1.0\linewidth}{!}{
\begin{tabular}{lcccccccccccccc}
\toprule
\multirow{2}{*}{Dataset} &\multirow{2}{*}{Method} & \multicolumn{2}{c}{Generation} & \multicolumn{5}{c}{Depth} & \multicolumn{5}{c}{Optical Flow} \\
\cmidrule(lr){3-4} \cmidrule(lr){5-9} \cmidrule(lr){10-14}
 ~ & ~& FID $\downarrow$ & FVD $\downarrow$ & 
 AbsRel $\downarrow$ & $\delta_1 \uparrow$ & $\delta_2 \uparrow$ & $\delta_3 \uparrow$ & RMSE $\downarrow$ & 
 EPE $\downarrow$ & AE $\downarrow$ & 1PE $\downarrow$ & 2PE $\downarrow$ & 3PE $\downarrow$\\
\midrule

\multirow{6}{*}{DSEC} & TAU~\cite{TAU} & 126.5 & 1052.9 & \multicolumn{10}{c}{Unsupported} \\
 ~ & FLODCAST~\cite{FLODCAST} & \multicolumn{2}{c}{Unsupported} & 0.54 & 26.5 & 63.2 & 74.6 & 10.3 & 15.3 & 25.8 & - & - & - \\
 ~ & FUTURIST~\cite{FUTURIST} & \multicolumn{2}{c}{Unsupported} & 0.40 & 38.1 & 66.9 & 85.8 & \textbf{8.0} & \multicolumn{5}{c}{Unsupported} \\
 ~ & UniFuture-Depth & \textbf{46.2} & 285.7 & 0.35 & 39.7 & 71.2 & 89.5 & 8.9 & \multicolumn{5}{c}{Unsupported} \\
 ~ & UniFuture-Flow & 54.3 & 314.8 & \multicolumn{5}{c}{Unsupported} & 12.4 & 32.4 & 96.3 & 88.1 & 79.1\\
\cmidrule(lr){2-14}
 ~ & \textbf{UniDynamics} & 50.9 & \textbf{224.0} & \textbf{0.30} & \textbf{47.9} & \textbf{78.1} & \textbf{92.9} & {8.1} & \textbf{6.8} & \textbf{23.4} & \textbf{93.9} & \textbf{80.0} & \textbf{65.6} \\

\midrule
\multirow{4}{*}{MVSEC} & TAU~\cite{TAU} & 136.9 & 1335.5 & \multicolumn{10}{c}{Unsupported} \\
~ & UniFuture-Depth & \textbf{69.8} & 694.8 & 0.23 & 34.3 & \textbf{92.7} & 99.8 & - & \multicolumn{5}{c}{Unsupported} \\
 ~ & UniFuture-Flow & 88.9 & 722.7 & \multicolumn{5}{c}{Unsupported} & 17.9 & 51.9 & 90.9 & 83.2 & 76.7\\
\cmidrule(lr){2-14}
 ~ & \textbf{UniDynamics} & 77.2 & \textbf{427.0} & \textbf{0.21} & \textbf{48.8} & {91.3} & \textbf{99.9} & {-} & \textbf{11.5} & \textbf{45.2} & \textbf{86.8} & \textbf{75.8} & \textbf{67.5} \\
 
\midrule
\multirow{4}{*}{M3ED} & TAU~\cite{TAU} & 156.1 & 1047.9 & \multicolumn{10}{c}{Unsupported} \\
~ & UniFuture-Depth & {88.6} & 705.4 & 0.23 & 64.4 & 86.5 & 94.1 & 10.3 & \multicolumn{5}{c}{\textbf{Unsupported}} \\
 ~ & UniFuture-Flow & 88.1 & 650.9 & \multicolumn{5}{c}{Unsupported} & - & - & - & - & -\\
\cmidrule(lr){2-14}
 ~ & \textbf{UniDynamics} & \textbf{68.5} & \textbf{508.3} & \textbf{0.22} & \textbf{66.5} & \textbf{87.3} & \textbf{94.5} & \textbf{10.1} & {-} & {-} & {-} & {-} & {-} \\

\bottomrule
\end{tabular}
}
\end{table*}

\noindent \textbf{Zero-shot on real-world datasets.}
We evaluate cross-domain and sim-to-real generalization by directly transferring the model pretrained on E-VKitti2 to three real-world datasets. As shown in~\cref{tab:dsec_zeroshot}, although our FID is slightly inferior to UniFuture-Depth on DSEC and MVSEC, the introduction of events and optical flow substantially improves FVD performance. Moreover, our method achieves clearly better generation quality on the more complex M3ED dataset, indicating more temporally coherent and physically plausible scene evolution. In addition, our perception results improve substantially, demonstrating the effectiveness of PDS in modeling geometry and motion cues. Overall, \textbf{UniDynamics} generalizes well across domains and produces more coherent and accurate future 4D scene forecasts in real scenes.

\subsection{Ablation Study}
We conduct ablation studies on E-VKitti2~\cite{VKitti2} to verify the contribution of the key components of UniDynamics, and report the results in~\cref{tab:ablation}.

\noindent \textbf{Event Inputs.} 
We remove the event input (\textit{UniDynamics (w/o events)}) and condition the model only on a single RGB frame. The results show a substantial drop in generation quality and coherence (FID increases from 39.1 to 46.8, and FVD from 223.6 to 577.1), along with large degradations in depth and optical-flow accuracy. This confirms that events provide crucial motion cues and rich texture/structure information for guiding future scene evolution as well as geometry and motion modeling. Moreover, removing the ELE module, i.e., the ``hard in'' setting, and directly concatenating event features to the noisy latent variables of future frames as RGB features leads to performance degradation. This suggests that ELE effectively inject the useful motion cues and texture information of events to the diffusion process while preserving the inherent static appearance reference from RGB.

\noindent \textbf{Flow Branch.} 
We remove the optical-flow branch (\textit{UniDynamics-Depth}) and keep only the depth branch. Without flow, PDS no longer needs to account for flow-related gradient constraints, and since depth and RGB are both static scene descriptors, the depth branch can focus more on geometric regression, making a slight improvement reasonable. However, the model simultaneously loses an explicit motion-field constraint, so the diffusion process must infer complex long-horizon dynamics mainly from single-frame appearance and local event cues. As a result, spatiotemporal consistency and physically plausible motion propagation become harder to maintain, leading to degraded RGB FID/FVD.

\noindent \textbf{Depth Branch.}
Removing the depth branch (\textit{UniDynamics-Flow}) leads to decreases in both RGB quality (FID/FVD) and optical-flow accuracy, due to the depth latent being injected as a geometry prior into the U-Net’s spatial layers, and plays a crucial role in stabilizing scene structure and boundaries throughout the diffusion process. Moreover, reliable optical flow estimation also relies on stable static structure and well-aligned motion boundaries. When geometric constraints are absent, the motion branch must estimate flow solely from the RGB latent, leading to higher EPE/AE and worse nPE metrics. Overall, depth and flow are complementary: depth stabilizes spatial structure, which benefits both appearance synthesis and motion estimation.

\noindent \textbf{Task Adapters in PDS.}
We further remove the task adapters and the gated bidirectional interaction in PDS to verify the effectiveness of task decoupling in PDS, and instead perform 3D convolutions directly in the joint layer to jointly model depth and optical flow. The resulting unified feature is then injected into the U-Net decoder via zero-conv. The results show clear degradations in generation quality and coherence, as well as in the perception metrics. This is because jointly modeling geometry and motion cues within a joint space tends to induce a gradient ``tug-of-war'', and injecting such entangled features can disturb the static appearance representations. In contrast, the task adapters in PDS explicitly decouple depth and flow to alleviate gradient conflicts, while the gated bidirectional interaction enables necessary information exchange without sacrificing task-specific modeling. Moreover, the two features are injected into the spatial and temporal layers according to their physical roles, further improving the accuracy and spatiotemporal consistency of 4D scene generation.

\begin{table*}[t]
\caption{Ablation studies of four key components  on E-VKitti2~\cite{VKitti2}.}
\label{tab:ablation}
\centering
\resizebox{1.0\linewidth}{!}{
\begin{tabular}{lcccccccccccc} 
\toprule
\multirow{2}{*}{Setting} & \multicolumn{2}{c}{Generation} & \multicolumn{5}{c}{Depth} & \multicolumn{5}{c}{Optical Flow} \\
\cmidrule(lr){2-3} \cmidrule(lr){4-8} \cmidrule(lr){9-13}
 & FID $\downarrow$ & FVD $\downarrow$ & 
 AbsRel $\downarrow$ & $\delta_1 \uparrow$ & $\delta_2 \uparrow$ & $\delta_3 \uparrow$ & RMSE $\downarrow$ &  
 EPE $\downarrow$ & AE $\downarrow$ & 1PE $\downarrow$ & 2PE $\downarrow$ & 3PE $\downarrow$ \\ 

\midrule
UniDynamics (w/o events) & 46.8 & 577.1 & 0.36 & 56.0 & 83.9 & 90.3 & 59.0 & 5.32 & 34.1 & 79.2 & 61.5 & 49.1 \\

UniDynamics (hard in) & 43.3 & 417.6 & 0.33 & 60.2 & 85.8 & 91.4 & 56.7 & 4.63 & 28.2 & 71.8 & 53.5 & 41.4 \\

{UniDynamics-Depth} (w/o flow) & 42.0 & {264.2} & \textbf{0.27} & \textbf{68.6} & \textbf{88.5} & \textbf{93.0} & \textbf{53.9} & \multicolumn{5}{c}{Unsupported} \\

{UniDynamics-Flow} (w/o depth) & 44.1 & {268.3} & \multicolumn{5}{c}{Unsupported} & {3.64} & {24.8} & {64.4} & {45.3} & {34.1} \\

{UniDynamics} (w/o adapters) & 50.8 & 384.9 & 0.30 & 62.6 & 85.6 & 91.6 & 56.0 & 4.60 & 30.3 & 72.3 & 53.5 & 41.9  \\

\midrule
\textbf{UniDynamics} (ours) & \textbf{39.1} & \textbf{223.6} & 
\textbf{0.27} & {67.1} & {87.8} & {92.7} & {54.7} & 
\textbf{3.00} & \textbf{21.0} & \textbf{58.1} & \textbf{38.5} & \textbf{27.7} \\
\bottomrule
\end{tabular}
}
\end{table*}

\section{Conclusion}
Toward practical deployment, we propose \textbf{UniDynamics}, a diffusion model conditioned on an event–RGB pair to generate future RGB, depth, and optical flow. We introduce event streams into the diffusion process via an Event Latent Enhancement (ELE) module, using them as complementary cues to RGB that provide effective initial momentum and reliable appearance guidance for scene extrapolation. We further decouple depth and flow in a Perceptual Dynamics Space (PDS) and tightly couple it with the U-Net denoising process to enforce physically plausible and spatiotemporally coherent 4D scene evolution. Experiments on both synthetic and real datasets demonstrate high-quality, temporally coherent appearance prediction and accurate geometric and motion perception.


\section*{Acknowledgements}
This work was supported by the National Natural Science Foundation of China (Grant No. 62273093, 62236002).

%
%
\bibliographystyle{splncs04}
\bibliography{main}

\end{document}